\pdfoutput=1
\documentclass[runningheads]{llncs}
\usepackage{graphicx}
\usepackage{amsmath} 
\usepackage[caption=false]{subfig} 
\usepackage{url}
\begin{document}
\title{Identification of Prototypical Task Executions Based on Smoothness as Basis of Human-to-Robot Kinematic Skill Transfer}

%
\titlerunning{Smoothness-Based Kinematic Skill Transfer}
%
\author{Jaime L. Maldonado C.\orcidID{0000-0002-2334-1073} \and
Christoph Zetzsche 
}
\authorrunning{J. Maldonado and C. Zetzsche}
%
\institute{Cognitive Neuroinformatics, University of Bremen, Bremen, Germany 
\email{jmaldonado@uni-bremen.de, zetzsche@informatik.uni-bremen.de} \\
}
\maketitle              

\begin{abstract}
In this paper we investigate human-to-robot skill transfer based on the identification of prototypical task executions by clustering a set of examples performed by human demonstrators, where smoothness and kinematic features represent skill and task performance, respectively. We exemplify our skill transfer approach with data from an experimental task in which a tool touches a support surface with a target velocity. Prototypical task executions are identified and transferred to a generic robot arm in simulation. The results illustrate how task models based on skill and performance features can provide analysis and design criteria for robotic applications.

\end{abstract}

\section{Introduction}
Modeling motor control during task execution is relevant for both behavioral research and robot applications aiming to achieve human-like performance. In this paper we investigate human-to-robot skill transfer based on the identification of prototypical task executions from a set of examples performed by human demonstrators. We approach the skill transfer from human to robot as an end effector control problem. For this purpose, we identify prototypical task executions by clustering demonstrations based on kinematic and smoothness features. In our approach smoothness features model the skill with which a task is executed and kinematic features represent performance criteria. 

As a use case of skill transfer we demonstrate our approach by extracting prototypical task executions from an experiment executed in virtual reality \cite{Maldonado_etal_2019}. In the experiment participants were requested to touch a box with tip of a spoon with a target contact velocity by performing a vertical planar movement. Each trial was represented by a smoothness feature, which quantifies a subject's control ability, and a kinematic feature, which quantifies task performance. Feature tuples were clustered using the K-means clustering algorithm which assigns data to groups and computes the cluster centers. The cluster centers are used to retrieve the most representative (i.e. prototypical) task executions from the data set. We exemplify the extent to which a robot can replicate the identified task execution prototypes by computing the joint velocities that are needed to achieve a target end-effector Cartesian velocity in a generic 6-DOF robot arm model. In this use case we illustrate how modeling task execution by means of skill and performance features provides a convenient representation which associates human and robot task performance for the analysis and design of robotic systems.

\section{Related Work} \label{sec:related_work}
\subsection{Speed Profiles of Arm and Hand Actions}\label{sec:velocity_profiles}
Bell-shaped speed profiles have been observed in point-to-point arm movements executed in different conditions (e.g. \cite{Nelson_1983,Nagasaki_1989,Schaal_2002}). The bell shape of the speed profiles of arm movements is regarded as a robust characteristic of planar and spatial movement\cite{Ambike_Schmiedeler_2013}. Discrete arm movements of healthy subjects show stereotypical smooth kinematics observed as a single peak in the speed profile \cite{Balasubramanian_etal_2015}. The rate at which the direction of movement can be changed depends on the physical characteristics of the arm and/or the coupling of the hand and the manipulated object. For example, bell-shape profiles with different lengths of the acceleration and deceleration phases have been reported for up and down movements under different speed and load conditions \cite{Papaxanthis_etal_1998}.

\subsection{Movement Smoothness}
Smoothness has been regarded as the organizing principle of motion \cite{Hogan_1984} and as a characteristic of skilled movements \cite{Nagasaki_1989}. Smooth movements are those in which the accelerative transients are kept to a minimum \cite{Hogan_1984}, thus in the context of sensorimotor control smoothness is related to the continuality or non-intermittency of a movement \cite{Balasubramanian_etal_2015}. Balasubramanian \textit{et al}.\ \cite{Balasubramanian_etal_2015} state that smoothness measures can be used to infer the control ability of a subject, since factors like poor motor planning/execution or familiarity with the task, among others, can lead to intermittencies. Regarding the use of smoothness to measure the a subject's ability it is important to note that smoothness is strongly task-dependent \cite{Balasubramanian_etal_2015}.

The spectral arc length (SPARC) quantifies smoothness independent of the movement's duration or amplitude, with larger values corresponding to smoother movements.  SPARC, defined in equation \ref{eq:SPARC}, is based on the computation of the arc length of the Fourier magnitude spectrum within the frequency range 0 to 20 Hz of a speed profile $v(t)$ \cite{Balasubramanian_etal_2015}: 

\begin{equation} \label{eq:SPARC}
\begin{gathered}
SPARC = - \int_{0}^{\omega _c}\left [ \left ( \frac{1}{\omega _c} \right )^2 + \left ( \frac{d\hat{V}(\omega))}{d\omega} \right )^2 \right ]^2 d\omega
\\
\hat{V}(\omega)=\frac{V(\omega)}{V(0)}
\end{gathered}
\end{equation}

where $V(\omega)$ is the Fourier magnitude spectrum of $v(t)$, $\hat{V}(\omega)$ is the normalized magnitude spectrum with respect to the DC magnitude $V(0)$ and $\omega_c$ is adaptively selected (for further details see \cite{Balasubramanian_etal_2015}).

\section{Identification of Prototypical Task Executions Based on Smoothness Features}
In the analysis of human movement and task executions kinematic features are typically used to describe and compare movement sequences or trajectory segments. Kinematic features can be used to quantify the between- and within-subjects variability of movement both when the task is fixed or across different tasks. Even if the task is specified with some performance criteria (e.g. to be executed within a certain time) variability might occur due to, for example, differences in body dimensions, skillfulness, or even fatigue when a movement is executed repeatedly. Since the task constraints determine the amount of intermittency in a movement and, thus, its smoothness, Balasubramanian \textit{et al}.\ \cite{Balasubramanian_etal_2015} suggest that smoothness can be used to distinguish control abilities in different subjects performing a particular task. Thus, kinematic features together with smoothness measures can be used to obtain models of the skill with which an action is executed.

In this paper we propose to model task execution by partitioning a set of demonstrations based on kinematic and smoothness features. Tuples of smoothness and kinematic features can be obtained for each example task execution. These tuples, which represent the control ability and a performance measure (e.g. maximum velocity or maximum acceleration) can be used to partition the data by assigning task executions to groups based on their similarity.

Clustering methods can be used to obtain models of task executions by identifying groups in the features tuples. For this purpose we use the K-means clustering algorithm. In addition to assign data to clusters, this unsupervised algorithm also identifies cluster centers which are representative of each group in the data. In our application, the cluster centers are interpreted as representative feature tuples which correspond to the prototypical task executions. Thus, the prototypes represent the different motor strategies with which the task was executed among a set containing within- and/or between-subjects variability. These prototypes can be used as basis for human-to-robot skill transfer.

\section{Human-to-Robot Kinematic Skill Transfer}\label{sec:skill_transfer}
Robot motion planning can be classified into four categories: 1) symbolic control, 2) end effector control, 3) joint level control, and 4) force/torque control \cite{Spiers_etal_2016}. In particular, end effector control refers to the placement of the end effector in Cartesian space. If data from human task execution in Cartesian space are available, the skill transfer from human to robot can be approached as an end effector control problem. In this case, for example, the hand is considered to be the human's end effector. 

Not all poses or motion trajectories are suitable of being replicated due to the mismatch between human and robot kinematics (e.g. in dimensions and degrees of freedom). The velocity of a robot's end effector depends on the velocities of the individual joints. Thus the extent to which a robot can replicate a human trajectory and the skill with which the movement is executed (encoded in the velocity and quantified by smoothness features) can be analyzed by computing the joint velocities that are needed to achieve a target end-effector Cartesian velocity. The computed joint velocities can be used to determine whether the execution of the movement will exceed the specifications of the actuators (e.g. maximum velocity). If the joint velocities are within the robot's specifications then a human-to-robot kinematic skill transfer can take place.   

The joint velocities can be computed by means of the manipulator Jacobian matrix $J$, which can be used to relate the configuration joint rate of change and the end-effector velocity as expressed in equation \ref{eq:jacobian} \cite{Corke_2017}:

\begin{equation} \label{eq:jacobian}
\dot{q}=J(q)^{-1}v
\end{equation}

where $q$ is the vector of joint angles and $v= (v_x, v_y, v_z, \omega_x, \omega_y, \omega_z)$ is the spatial velocity vector of the end effector which contains the rate of change of the end effector pose (i.e. the translational and rotational velocity) in the world frame, given that $J(q)$ is square and not singular. A square Jacobian matrix can be obtained for a robot with 6 joints operating in a 3-dimensional task space \cite{Corke_2017}.

The joint velocities necessary to reproduce a prototypical task execution are computed as follows: 1) obtain a configuration $q$ by means of inverse kinematics (i.e. the joint angles that correspond to and end effector position), 2) compute the manipulator Jacobian, 3) check if the manipulator Jacobian is singular (i.e. $det(J(q))\neq 0$), and 4) compute the joint velocity $\dot{q}$ as expressed in equation \ref{eq:jacobian}. The joint velocities necessary to reproduce the task execution prototypes were computed using the functions provided in the \textit{Robotics Toolbox for MATLAB} (release 10) \cite{Corke_2017} for a generic 6-DOF robot arm (model \textit{Simple6}, shown in Figure \ref{fig:robot_joints}).

\section{Dataset}
\begin{figure}
	\subfloat[Experimental setup and virtual environment  \label{fig:exp-setup}]{%
		\includegraphics[width=0.28\textwidth,height=0.14\textheight]{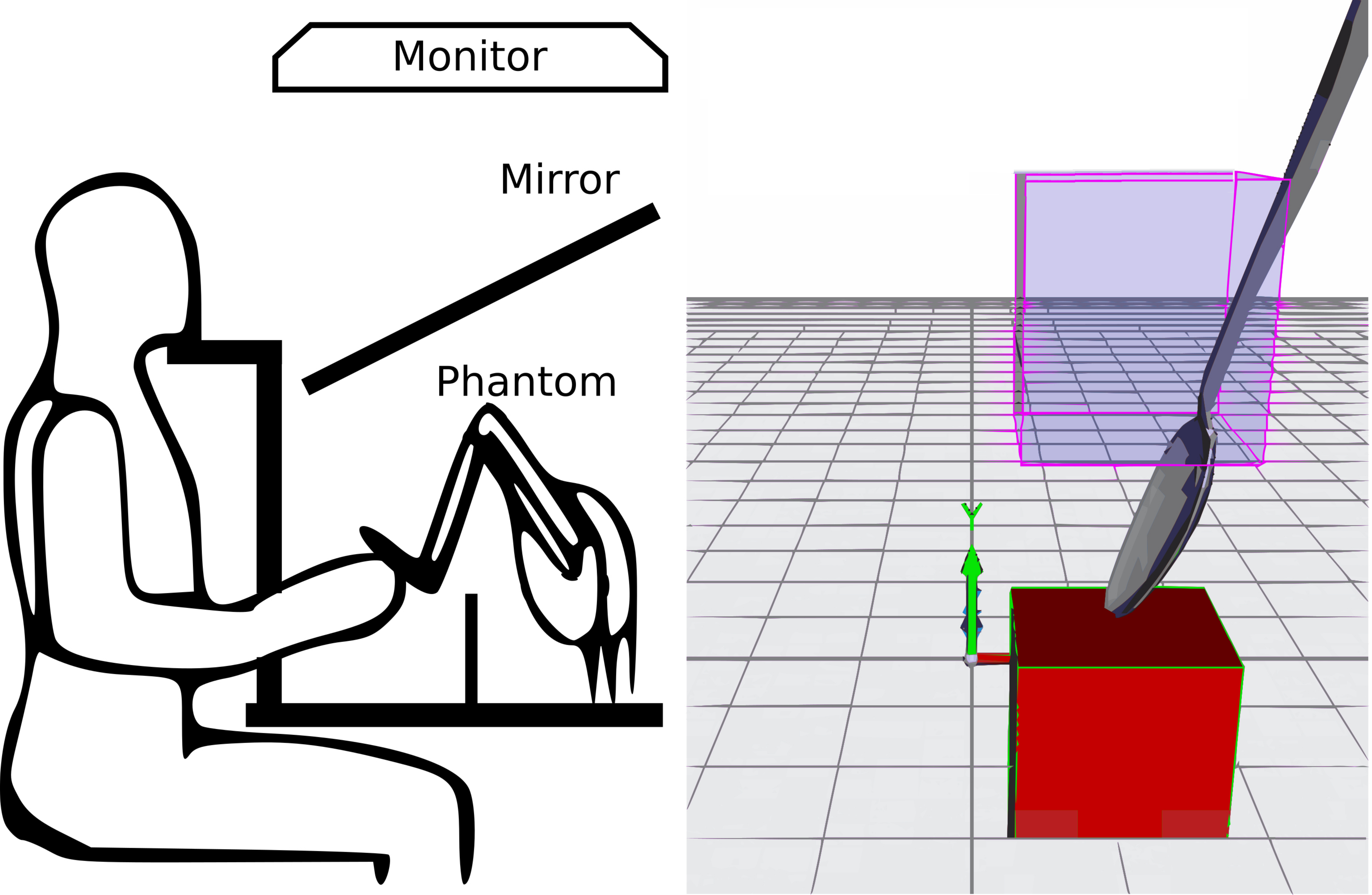}
	}\hfil
	\subfloat[Vertical trajectories and speed profiles for all participants profiles.\label{fig:position_speed_trajectories}]{%
		\includegraphics[width=0.65\textwidth]{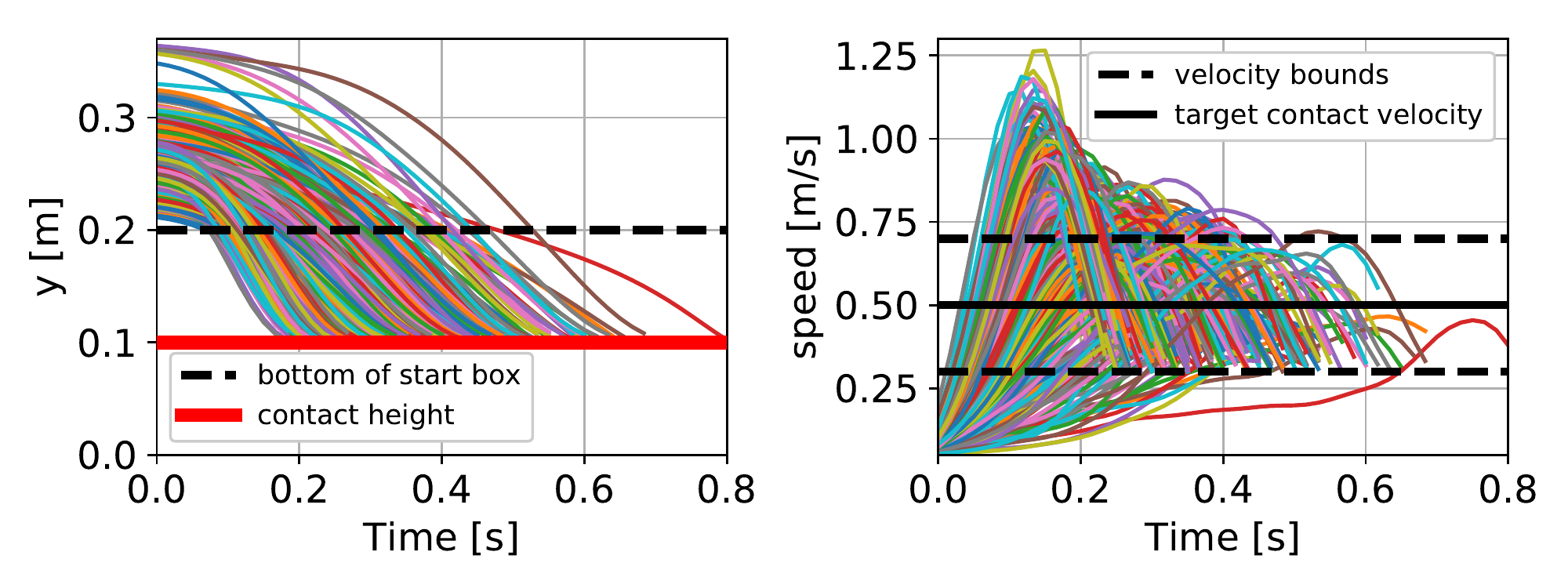}
	}
	
	\caption{Schematic illustration of the experimental setup and experimental data. The axes in \ref{fig:exp_setup_and_data} are labeled as $x$ mediolateral (side-to-side), $y$ vertical (up-down), and $z$ anteroposterior (forward-backward).}
	\label{fig:exp_setup_and_data}
\end{figure}

We demonstrate our skill transfer approach by identifying prototypical task executions from data acquired in a virtual reality experiment \cite{Maldonado_etal_2019}. The experimental setup is shown in Figure \ref{fig:exp-setup}. Participants viewed a 3D environment containing a solid red box (termed contact box), a semi-transparent cube (termed start box)  and a spoon. The virtual spoon was attached to the stylus of a Phantom Premium 1.5 (3D Systems) force-feedback device, enabling the rendering of contact force between the surface of the contact box and the tip of the spoon. Position data from the Phantom were sampled at 60 Hz. Ten volunteers participated in the experiment, each performing 150 trials. 

Participants were requested to touch the contact box with a target velocity of 0.5 $m/s$ by performing a planar vertical movement starting from the start box. We use all trials in which $0.3 m/s \leq contact\ velocity \leq 0.7m/s $ as examples of different demonstrators executing the task (650 trials in total). SPARC was computed for the trajectory segment from the start box until the spoon makes contact with the box. The onset of the movement was determined as the time instant in which $speed > 0.05 m/s$. Tuples of peak velocity magnitude (PV) and SPARC were clustered with the K-means algorithm in order to obtain prototypes of task performance. The trajectories and the speed profiles of all subjects are shown in Figure \ref{fig:position_speed_trajectories}. It is important to observe that, even for the simple task investigated here, the peak speed, contact speed and movement duration
show a large variability.

\section{Results}

\begin{figure}
	\subfloat[Cluster assignation and cluster centers obtained from PV-SPARC tuples.\label{fig:clustered_pv_sparc}]{%
		\includegraphics[width=0.48\textwidth]{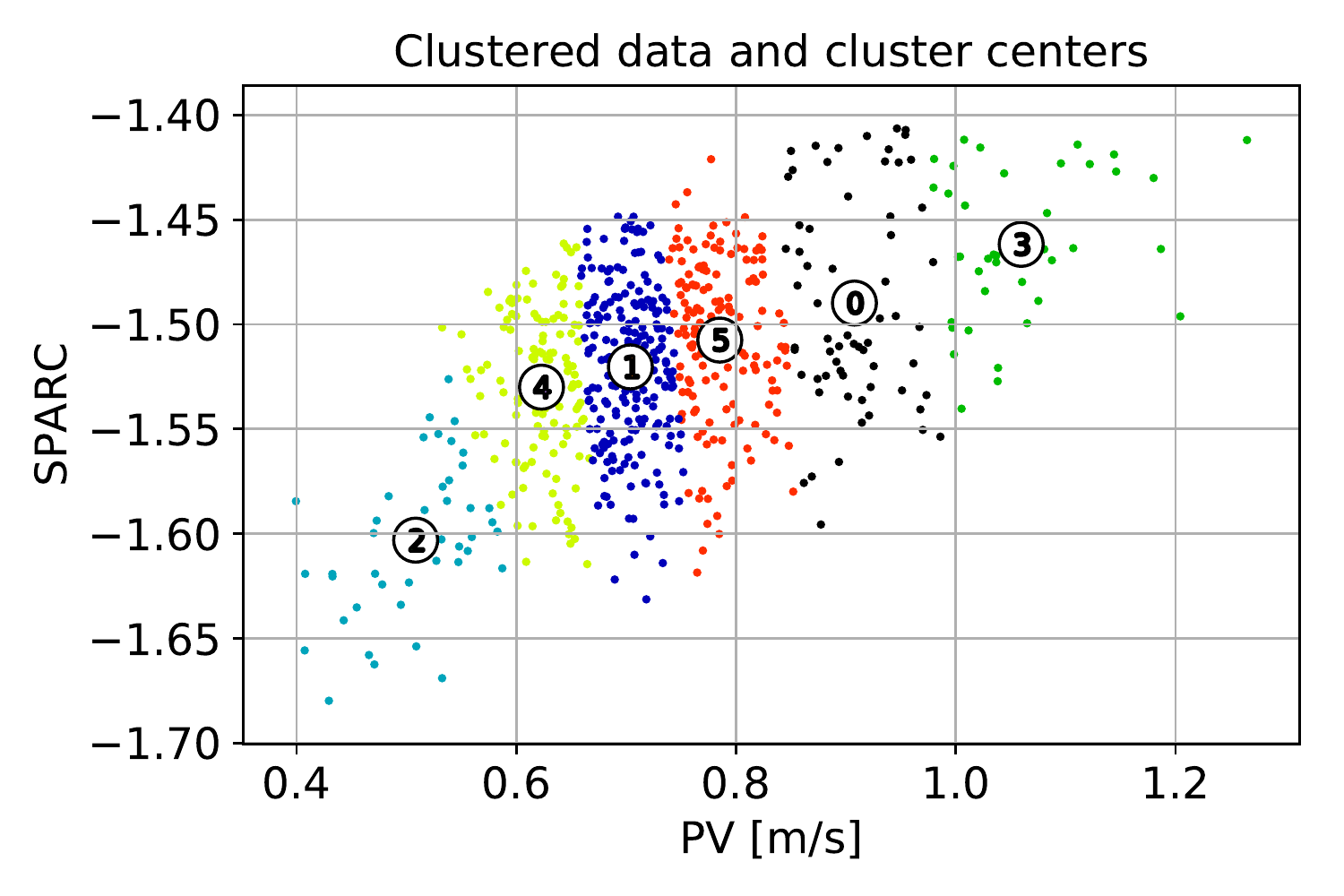}
	}\hfil
	\subfloat[Speed profiles of the prototypes obtained with the clustering of PV-SPARC tuples.\label{fig:pv_sparc_prototypes}]{%
		\includegraphics[width=0.48\textwidth]{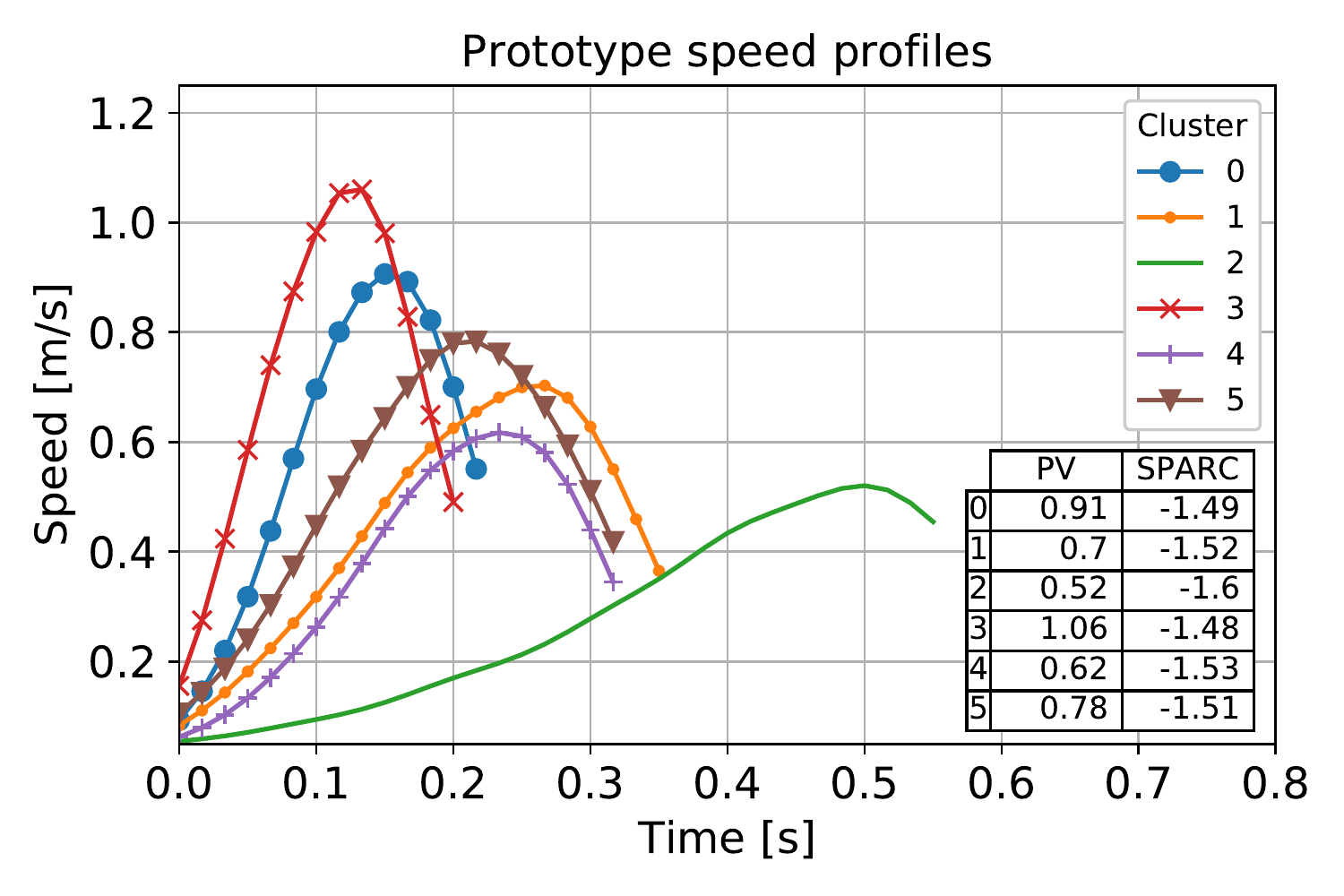}
	}
	
	\caption{Clustering of PV-SPARC tuples and speed profiles of the cluster prototypes}
	\label{fig:clustering_results}
\end{figure}

Silhouette analysis, which can be used to determine the number of clusters (for details see \cite{scikit_learn_cluster_selection}), reveals that $2 \leq n\ clusters \leq 16 $ produces valid clusters. For simplicity of interpretation, the PV-SPARC tuples were grouped into 6 clusters, which produced cluster centers with well separated PV values over the range of the experiment data (average silhouette score = 0.337). The K-means clustering and the silhouette analysis were computed with the \textit{scikit-learn} machine learning library \cite{scikit-learn}. The clusters and cluster centers are shown in Figure \ref{fig:clustered_pv_sparc}. The prototypical task executions were retrieved by identifying the closest tuple to each cluster center. The speed curves of the prototypes are shown in Figure \ref{fig:pv_sparc_prototypes} with their corresponding PV-SPARC values. 

\begin{figure}
	\centering
	\subfloat[Robot joint labels\label{fig:robot_joints}]{%
		\includegraphics[width=0.28\textwidth]{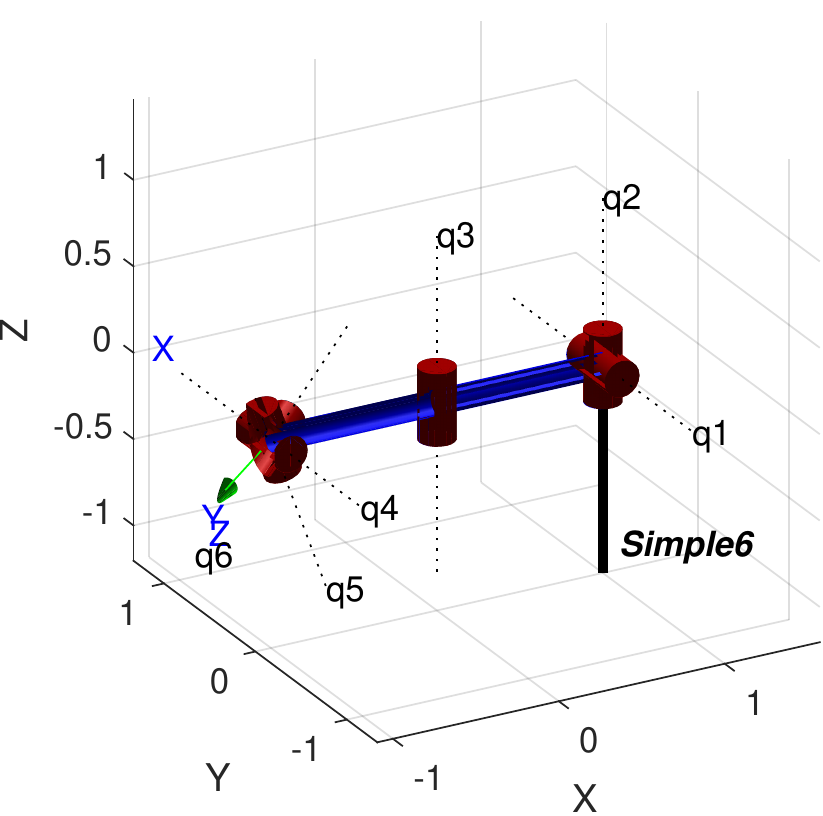}
	}\hfil
	\subfloat[Default robot pose \label{fig:robot_pose}]{%
		\includegraphics[width=0.28\textwidth]{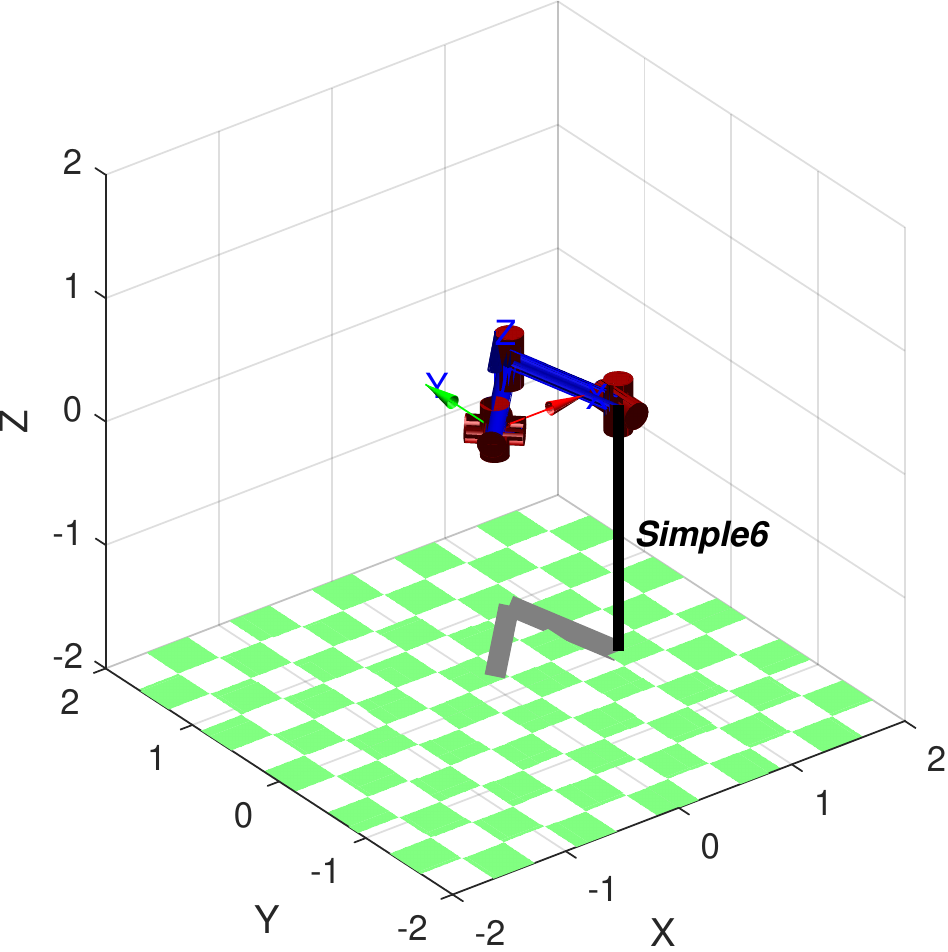}
	}
	
	\caption{\textit{Simple6} generic 6-DOF robot arm (available in the \textit{Robotics Toolbox for MATLAB} \cite{Corke_2017})}
	\label{fig:robot_joints_pose}
\end{figure}

\begin{figure}
	\includegraphics[width=\textwidth]{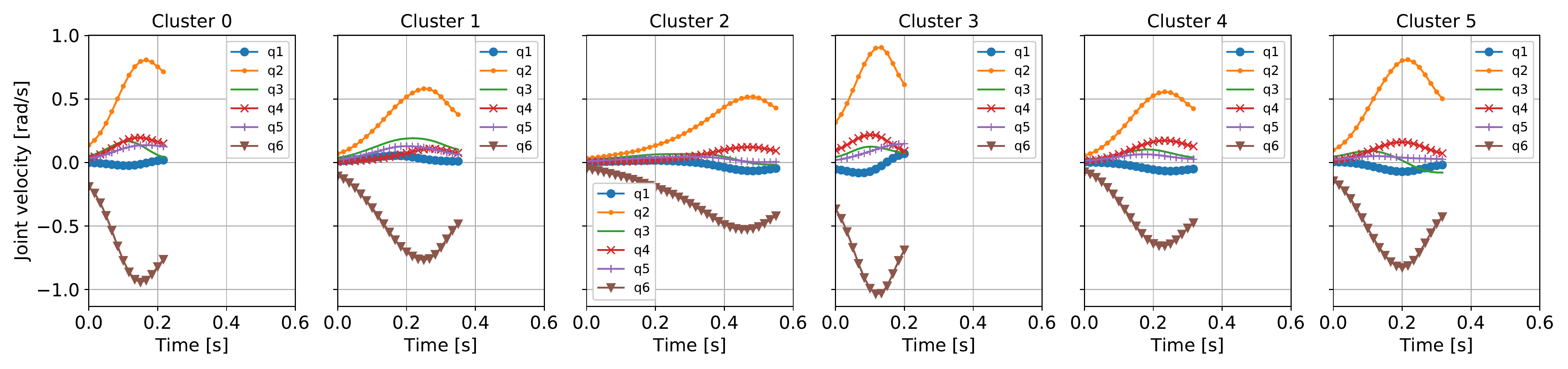}
	\caption{Joint velocities necessary to reproduce task executions prototypes identified in each cluster.} \label{fig:joint_vels_for_each_cluster}
\end{figure}

The \textit{Simple6} generic robot arm and its joint labels used to illustrate the process of kinematic skill transfer are shown in Figure \ref{fig:robot_joints}. The robot arm was set to a default dexterous configuration (q1=0 deg, q2=-60 deg, q3=120 deg, q4=90 deg, q5=-90 deg and q6=30 deg; see Figure \ref{fig:robot_pose}). The joint velocities for each task prototype, computed for the \textit{Simple6} arm following the steps described in section \ref{sec:skill_transfer}, are shown in Figure \ref{fig:joint_vels_for_each_cluster}. Note that movement time corresponds to that of the prototypes shown in Figure \ref{fig:pv_sparc_prototypes} and that the highest joint velocities are observed for the reproduction of the prototypes from clusters 0 and 3, which correspond to the prototypes with the largest peak speeds.  


\section{Conclusion and Discussion}
In this paper we introduced an approach to identify prototypical task executions from a set of examples performed by human demonstrators. Clustering feature tuples can be used in applications where a performance criterion is given,  for example, expressed as a smoothness and peak speed requirement. 

In robotics two design factors must be addressed: 1) which aspect of human motor control can be implemented and 2) the extent to which human performance parameters (e.g. maximum acceleration) match those of the robot actuator. Our approach can provide information for both design factors. Given that the retrieved task performance prototypes can be realized with the robot actuators, movement or speed profiles can be implemented in a robot system by passing them to the robot's end-effector controller. Otherwise, information of the parameters of the prototypes exceeding the robot's specifications can be used to scale the task parameters accordingly.

\section*{Acknowledgements}
The research reported in this paper has been supported by the German Research Foundation, as part of Collaborative Research Center 1320 "EASE - Everyday Activity Science and Engineering", University of Bremen (http://www.ease-crc.org/). The research was conducted in subproject H01 - Acquiring activity models by situating people in virtual environments.
%
%
%
%
%
%
\bibliographystyle{splncs04}
\bibliography{ki_2020_refs}
\end{document}